\pdfoutput=1
\documentclass[11pt]{article}

\usepackage[table]{xcolor}

\usepackage[final]{acl}

\usepackage{times}
\usepackage{latexsym}

\usepackage[T1]{fontenc}
\usepackage[utf8]{inputenc}

\usepackage{microtype}

\usepackage{inconsolata}

\usepackage{graphicx}

\usepackage{multirow}
\usepackage{array}
\newcommand{\ph}[1]{\phantom{#1}}
\usepackage{quoting}
\quotingsetup{leftmargin=10pt, rightmargin=0pt}
\usepackage{caption} 

\title{An LLM Benchmark for Addressee Recognition\\in Multi-modal Multi-party Dialogue}

\author{
 \textbf{Koji Inoue},
 \textbf{Divesh Lala},
 \textbf{Mikey Elmers},
 \textbf{Keiko Ochi},
 \textbf{Tatsuya Kawahara}
\\
\\
 Graduate School of Informatics, Kyoto University, Japan, \\
\\
 \small{
   \textbf{Correspondence:} \href{mailto:inoue@sap.ist.i.kyoto-u.ac.jp}{inoue@sap.ist.i.kyoto-u.ac.jp}
 }
}

\begin{document}
\maketitle
\begin{abstract}
Handling multi-party dialogues represents a significant step for advancing spoken dialogue systems, necessitating the development of tasks specific to multi-party interactions.
To address this challenge, we are constructing a multi-modal multi-party dialogue corpus of triadic (three-participant) discussions.
This paper focuses on the task of addressee recognition, identifying who is being addressed to take the next turn, a critical component unique to multi-party dialogue systems.
A subset of the corpus was annotated with addressee information, revealing that explicit addressees are indicated in approximately 20\% of conversational turns.
To evaluate the task's complexity, we benchmarked the performance of a large language model (GPT-4o) on addressee recognition.
The results showed that GPT-4o achieved an accuracy only marginally above chance, underscoring the challenges of addressee recognition in multi-party dialogue. 
These findings highlight the need for further research to enhance the capabilities of large language models in understanding and navigating the intricacies of multi-party conversational dynamics.
\end{abstract}

\renewcommand{\thefootnote}{\fnsymbol{footnote}}
\footnote[0]{This paper has been accepted for presentation at International Workshop on Spoken Dialogue Systems Technology 2025 (IWSDS 2025) and represents the author’s version of the work.}
\renewcommand{\thefootnote}{\arabic{footnote}}

\section{Introduction}

The rapid advancements in dialogue systems, fueled by the emergence of large language models (LLMs) capable of generating human-like text and engaging in natural conversations, have been largely confined to the realm of dyadic interactions.
While these systems have demonstrated remarkable progress, they fail to capture the complexities inherent in multi-party dialogues, involving three or more participants.
These dialogues are characterized by intricate information flow, dynamic participant roles, and nuanced social cues, posing significant challenges for system development.

Previous research has explored specific aspects of multi-party dialogues, including turn-taking~\cite{Lee2024, Auer2018, Skantze2015}, addressee recognition~\cite{le2019speaking, li2023pre, tan2023chatgpt}, and dialog act recognition \cite{qamar2023speaking}.
However, existing benchmarks are limited by their reliance on text-based or acted dialogue data, failing to reflect the spontaneity and multi-modality inherent in natural human interactions.

\begin{figure}[t]
    \centering
    \includegraphics[width=1\linewidth]{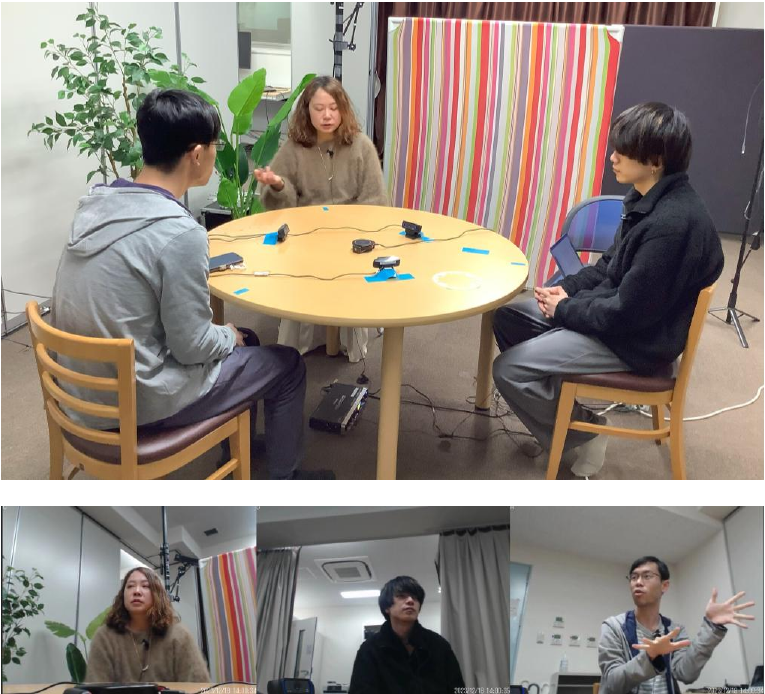}
    \caption{A snapshot from TEIDAN corpus}
    \label{fig:teidan}
\end{figure}

\begin{table*}[t]
    \centering
    \caption{Statistics of turn and addressee annotation}
    \begin{tabular}{cccccc}
    \hline
    Session ID & Time & \# IPU (A/B/C) & \# Turn (A/B/C) & \# Addressed & \# Not Addressed \\
    \hline
    session-01-city & 6:14 & \ph{0}65 / \ph{0}81 / 137 & 12 / 16 / 13 & 9 & 32 \\
    session-02-city & 5:50 & \ph{0}76 / \ph{0}94 / \ph{0}98 & 16 / 22 / 23 & 14 & 47 \\
    session-03-city & 6:12 & \ph{0}81 / 123 / 128 & 19 / 29 / 30 & 6 & 72 \\
    session-04-city & 5:46 & 146 / 142 / 123 & 29 / 21 / 24 & 10 & 64 \\
    session-05-city & 5:18 & 108 / 119 / \ph{0}95 & 44 / 43 / 46 & 36 & 97 \\
    \hline
    Total (Ave.) & 29:20 (5:52) & 1616 (323.2) & 387 (77.4) & 75 (15) & 312 (62.4) \\
    \hline
    \end{tabular}
    \label{tab:annotation:stats}
\end{table*}

To address this crucial gap, this paper introduces a novel, spontaneous, and multi-modal multi-party dialogue corpus specifically designed to facilitate research on triadic (three-participant) dialogue systems. 
This research further focuses on the critical, yet under-explored, task of addressee recognition – the identification of the intended recipient of a turn – which is foundational for enabling dialogue systems to navigate and participate effectively in multi-party settings.
Unlike dyadic interactions where the addressee is implicitly defined, turn-taking in multi-party settings is far more complex.
The intended recipient might be a specific participant or the group as a whole, and behavioral signals are often subtle and inconsistent \cite{Auer2018, Skantze2015}.

This work makes two key contributions:
\begin{itemize}
    \setlength{\itemsep}{0mm}
    \setlength{\parskip}{0mm}
    \item The introduction of the TEIDAN corpus, a new dataset of spontaneous, multi-modal, triadic dialogues that provides a unique resource for this understudied area
    \item The development of the first LLM benchmark specifically designed for addressee recognition in multi-modal, multi-party dialogue, underscoring the challenges and the need for further innovation
\end{itemize}
Ultimately, this research aims to establish a strong foundation for the development of advanced multi-party dialogue systems capable of understanding and responding effectively in complex, real-world conversational settings.

\section{TEIDAN Corpus}

We begin by briefly describing the TEIDAN multi-party corpus.
Unlike other datasets that involve specific contexts, such as meetings \cite{Carletta2007,Mostefa2007}, task-oriented interactions \cite{Kontogiorgos2018,Nihei2014}, or game-based scenarios \cite{Stefanov2016,Litman2016,Hung2010}, the TEIDAN corpus captures goal-free discussions.
Additional multi-party corpora also exist for online discussions \cite{Reverdy2022}.

The discussions involved triads (groups of three participants).
Each participant was seated in a circle around a table placed at the center, as shown in Figure~\ref{fig:teidan}.
Cameras recorded each participant's face, and individual pin microphones were used to capture their speech separately.

Participants discussed three general topics: (1) which city would be best suited as an alternative capital of Japan, (2) which items would be essential to bring to a desert island, and (3) where they would like to travel on the weekend.
Each triad completed three discussion sessions, one for each topic.

Each triad conversed for approximately 5 to 10 minutes per session, with no requirement to reach a conclusion.
Data were collected from 10 triads, resulting in a total of 30 discussion sessions.
Note that this corpus is in the Japanese language.

\section{Annotation of Addressee}

We annotated a subset of the TEIDAN corpus for addressee information.
The annotation process consisted of the following steps:

(1) Initially, turns and the current speaker were annotated. Since the original TEIDAN corpus contains only IPU (inter-pausal unit) utterance segments, turn segments within the dialogue were annotated by removing certain utterances, including backchannels.
This ensured that only one speaker could hold the floor at any given time.
Minimal overlap was permitted during turn transitions.

(2) Following turn annotation, we labeled the addressee information to indicate whether the next speaker was explicitly addressed.
If addressed, the label corresponded to one of the participant IDs (e.g., A, B, or C).
Otherwise, it was labeled as `O', signifying that no specific individual was addressed and any participant could take the turn. 

This labeling process considered both textual and visual cues, such as gaze behavior.
Initially, a single session was annotated and discussed to ensure inter-rater agreement by the authors.
Subsequently, the remaining four sessions were annotated by one of the authors.

We have so far annotated five sessions from the TEIDAN corpus.
Table~\ref{tab:annotation:stats} summarizes the annotation results.  The analysis revealed that only approximately 20\% (75 / 387) of turns explicitly specify an addressee.
The `O' label, indicating no specific addressee, was prevalent, particularly in discussions involving multiple opinions (statements).
This result implies that a multi-party dialogue system that participates in this type of discussion and disregards addressee information may potentially interrupt the dialogue in 20\% of turn-taking instances, assuming that they can always correctly recognize the end of the turn of a human participant.

\begin{table}[t]
    \centering
    \caption{Performance of addressee recognition by GPT-4o}
    \begin{tabular}{cccc}
        \hline
        \multicolumn{1}{c}{LLM output} & \# Correct & \# Incorrect \\
        \hline
        Addressed (A/B/C) &  9 & 14 \\
        Not addressed (O) & 304 & 60 \\
    \hline
    \end{tabular}
    \label{tab:result:confusion}
\end{table}

\begin{table}[t]
    \centering
    \setlength{\tabcolsep}{1mm}
    \rowcolors{1}{white}{gray!20}
    \renewcommand{\arraystretch}{1.2}
    \caption{Context example where GPT-4o correctly recognized addressee as person C, translated from original Japanese utterances}
    \label{tab:result1}
    {\small
    \begin{tabular}{c m{70mm}}
    \hline
     & \multicolumn{1}{c}{Utterance} \\
    \hline
    C: & So, if we wanted to change the capital from Tokyo, where do you think would be a good place? \\
    A: & I think Osaka would be a good choice. Osaka is the largest city in western Japan, and in terms of population, there's no other city in western Japan that surpasses it. So, I think Osaka is a strong candidate. \\
    B: & But one of the reasons for wanting to relocate the capital from Tokyo is likely the population increase, or rather, Tokyo's population is becoming unmanageable, necessitating the transfer of some capital functions. (...) Hokkaido is a bit cold, though, so I think somewhere in Kyushu or, for example, the Tokai region might be better. \\
    A: & I see, that makes sense. \\
    B: & What do you think, Ochi-san? Do you have any specific ideas? \textbf{(addressee is C)} \\
    \hline
    \end{tabular}
    }
\end{table}

\section{Benchmark for Addressee Recognition}

To evaluate the task's complexity, we tested the performance of a multimodal large language model (GPT-4o) on addressee recognition.
The model was given a prompt as follows:
\begin{quote}
    In the following conversation among A, B, and C, please infer who is addressed as the next speaker in the last utterance.  
    Answer with one of the following: ``A, B, C, or O''.
    ``A, B, and C'' represent the participants, and ``O'' represents the case where no one is addressed, and anyone can take the turn next.
    The output should only contain the label ``A, B, C, or O'' and should not include any other characters.
\end{quote}
This was followed by five context turn utterances with the current utterance, and also it contained the name of the discussion topic and designated identifier of the participants.
% Then, we asked it to output an addressee label (A, B, C, or O). 
Note that the utterances were manually transcribed.

The GPT-4o achieved an accuracy of 80.9\%, which is only marginally above the chance level (80.6\%).
This indicates that the model struggles to identify the addressee in multi-party dialogues.
The output by the LLM is summarized in Table~\ref{tab:result:confusion} which shows that the model tends to output `O', indicating that it often fails to recognize when an utterance is directed at a specific participant.

We then analyzed samples to examine how GPT-4o deals with addressee recognition, as illustrated below:

\begin{table}[t]
    \centering
    \setlength{\tabcolsep}{1mm}
    \rowcolors{1}{white}{gray!20}
    \renewcommand{\arraystretch}{1.2}
    \caption{Context example where GPT-4o incorrectly recognized addressee as O, translated from original Japanese utterances}
    \label{tab:result2}
    {\small
    \begin{tabular}{c m{70mm}}
    \hline
     & \multicolumn{1}{c}{Utterance} \\
    \hline
    B: & A riddle. \\
    C: & When I suggested, it might be something related to Fukuoka, or perhaps Kitakyushu, this person insisted they were from Moji, mentioning some kind of ward distinction I didn't understand. So, I think in that sense, it's decentralized. \\
    B: & Hmm, it seems like the decentralization of cities is an unavoidable issue after all. \\
    C: & That's right. But Osaka has Umeda and... \\
    A: & Tennoji? \\
    C: & Not Tennoji, but Namba, I think. \textbf{(addressee is A)} \\
    \hline
    \end{tabular}
    }
\end{table}

\begin{table}[t]
    \centering
    \setlength{\tabcolsep}{1mm}
    \rowcolors{1}{white}{gray!20}
    \renewcommand{\arraystretch}{1.2}
    \caption{Context example where GPT-4o incorrectly recognized addressee as O, translated from original Japanese utterances}
    \label{tab:result3}
    {\small
    \begin{tabular}{c m{70mm}}
    \hline
     & \multicolumn{1}{c}{Utterance} \\
    \hline
    A: & One of the reasons why I prefer Osaka is that its city planning, including roads and railway networks, is very linear and easy to understand. \\
    C: & Like Midosuji? \\
    A: & Exactly. If you've ever seen a map of the Tokyo subway, you'll know that it's quite convoluted and complex. In contrast, Osaka's layout is more grid-like. \\
    C: & With streets like ``something-suji'' and ``Something-suji Line.'' \\
    A: & Yes. I think Tokyo is more circular, but a linear layout is easier to understand. Osaka's linear layout with clear divisions, like this area for administrative functions and this area as the central hub where people gather, makes it superior as a city, in my opinion. \\
    C: & I feel like in Nagoya, Sakae and Nagoya Station are slightly separated, aren't they? \textbf{(addressee is B)} \\
    \hline
    \end{tabular}
    }
\end{table}

\paragraph{(1) Explicit Question (Correct)}

An example in Table~\ref{tab:result1} shows a case where GPT-4o correctly identified the addressee as C because the final utterance, a question, was explicitly directed to that individual.
Although current LLMs effectively handle such explicit cases, the corpus contains many instances that are not as straightforward.

\paragraph{(2) False Negative}

In both examples presented in Table~\ref{tab:result2} and Table~\ref{tab:result3}, the GPT-4o's output indicated no specific addressee (O), while the reference labels were A and B, respectively.
This kind of false-negative instance represented the majority of errors in this experiment.
In the Table~\ref{tab:result2} example, the final speaker, C, was looking at person A, suggesting that gaze information is crucial for this task.
In the Table~\ref{tab:result3} example, the final speaker inquires about Nagoya, a city in Japan. 
Within the context of this discussion, B was about to recommend this city.
Therefore, this task also necessitates the consideration of such prior information.

% \paragraph{(3) False Positive}

% Table~\ref{tab:result4} reports another incorrect example where the MLLM output was C, but the reference was O.
% In this case, MLLM seemed to be biased to the previous C's utterance.

% \begin{table}[t]
%     \centering
%     \setlength{\tabcolsep}{1mm}
%     \rowcolors{1}{white}{gray!20}
%     \renewcommand{\arraystretch}{1.2}
%     \caption{Context example where GPT-4o incorrectly recognized addressee as person C (Reference is O), translated from original Japanese utterances}
%     \label{tab:result4}
%     {\small
%     \begin{tabular}{c m{70mm}}
%     \hline
%      & \multicolumn{1}{c}{Utterance} \\
%     \hline
%     B: & Right, right, right, right. It becomes the core, and it's hard to establish supply depots. \\
%     A: & If that's the case, Nagoya is in a pretty good position. \\
%     B: & Well, if we go by that argument, even if it's Nagoya, Osaka has a larger population and it's easier to relocate things there. (...) So, I think it would be easier to relocate government agencies to the Kinki region, and it would likely face less opposition from other areas. There's a sense of acceptance across Japan. \\
%     A: & But when you think about the industrial center, it's gotta be Toyota, you know. \\
%     C: & Yeah, yeah, Toyota, right, right, right. \\
%     A: & You know, with Toyota, their city is practically named after them. It's right there in the heart of such a major company. \\
%     \hline
%     \end{tabular}
%     }
% \end{table}

\section{Adding Gaze Features}

To see the effect of gaze information in the current task, we processed the video of each participant (shown at the bottom of Figure~\ref{fig:teidan}) and automatically annotated their gaze throughout the discussion.
OpenFace 2.0 was used to estimate the eye gaze vector \cite{Baltrusaitis2018, Wood2015}.
We could then generate a gaze vector 30 times a second.

For every gaze timestamp, we then estimated whether the gaze of the participant who had the turn (speaker) was directed at either one of the other participants or at nobody in particular.
As each participant was seated in an approximately equilateral triangle, we used a simple heuristic to test if the speaker was looking at another participant.
The y (up-down) portion of the gaze vector must be within a certain range (0.2), and the x (left-right) gaze vector had to be out of a certain range (-0.2 to 0.2).
If this heuristic was met, then the gaze timepoint was labeled as the speaker looking at the relevant participant, else the gaze timestamp was labeled as O (no participant).

\begin{table}[t]
    \centering
    \caption{Performance of addressee recognition by GPT-4o added simple gaze features}
    \begin{tabular}{cccc}
        \hline
        \multicolumn{1}{c}{LLM output} & \# Correct & \# Incorrect \\
        \hline
        Addressed (A/B/C) &  12 & 36 \\
        Not addressed (O) & 279 & 60 \\
    \hline
    \end{tabular}
    \label{tab:result:confusion:gaze}
\end{table}

 We also labeled the turn of a speaker as opposed to continuous timestamps.
 We based our approach on previous research which found that end-of-turn gaze was important \cite{kawahara2016multi,Degutyte2021} and labeled the \textit{majority} gaze in the turn's final second, specifically the gaze label which was present in over 50\% of the timestamps, or O if this was not reached.

 This information was added to the previous prompt to assess if adding gaze information in this way could improve the result.
 However, as shown in Table~\ref{tab:result:confusion:gaze}, adding this information did not improve accuracy, as the accuracy score (75.2\%) went down under the chance level.
 While future work necessitates manual annotation of gaze information, the current results indicate that existing LLMs also struggle to incorporate such additional modalities within the context of multi-party dialogues.

 % This was likely because the gaze information was not precise enough to be used in this way. Annotation was done automatically so it does not represent a completely accurate ground truth. This could only be done through manual annotation.

\section{Benchmark for Next Speaker Prediction}

We are also interested in predicting the actual next speaker in the multi-party scenarios~\cite{Lee2024,Lee2023}.
This is distinguished from addressee annotation, where subsequent information on who took the turn is unknown, and there is no `O' label as somebody must take the turn.
It is possible that the addressee and the actual next speaker differ because of interruptions during the turn.

We then evaluated GPT-4o's performance on this next speaker prediction task.
Using a prompt similar to that used for addressee recognition, the model was tasked with predicting the actual next speaker.
The output label was limited to A, B, or C, with a chance-level accuracy of 50\%, as either of the other two participants could take the turn.
As a result, GPT-4o attained an accuracy of 46.0\% on this task, performing below chance level.
This outcome further suggests that the model struggles to effectively capture the dynamics of turn-taking in spontaneous multi-party dialogues.

\section{Conclusion}

This study investigated the challenges of addressee and next speaker prediction in multi-party dialogues.
We introduced a new multi-party dialogue corpus and analyzed the performance of an LLM (GPT-4o) on these tasks.
The findings revealed that LLMs struggle with the complexities of multi-party interactions.
They perform only marginally above chance level in addressee recognition and below chance level in the next speaker prediction task.
Although the LLM was given the simple gaze feature, it did not improve the performance.

These results underscore the need for further research to improve LLMs' understanding of multi-party conversational dynamics.
Future work should explore more sophisticated methods for incorporating contextual information, including gaze and other non-verbal cues, and develop new models that can better capture the intricate interplay between participants in multi-party conversations.

\section*{Acknowledgments}

This work was supported by JST PREST JPMJPR24I4, JST Moonshot R\&D JPMJPS2011, and JSPS KAKENHI JP23K16901.
The authors also express appreciation to the members of speech and audio processing laboratory at Kyoto University for their participation in the data collection.

\bibliography{acl_latex}

% \appendix

% \section{Prompt} \label{sec:prompt}

% The prompt used in the addressee recognition task is as follow:
% \begin{quote}
%     In the following conversation among A, B, and C, please infer who is addressed as the next speaker in the last utterance.  
%     Answer with one of the following: ``A, B, C, or O''.
%     ``A, B, and C'' represent the participants, and ``O'' represents the case where no one is addressed, and anyone can take the turn next.
%     The output should only contain the label ``A, B, C, or O'' and should not include any other characters.
% \end{quote}
% Note that this is translated from the original Japanese one.

\end{document}